# Closing the Learning-Planning Loop with Predictive State Representations

Byron Boots
Machine Learning Department
Carnegie Mellon University
Pittsburgh, PA 15213
beb@cs.cmu.edu

Sajid M. Siddiqi
Robotics Institute
Carnegie Mellon University
Pittsburgh, PA 15213
siddiqi@cs.cmu.edu

Geoffrey J. Gordon
Machine Learning Department
Carnegie Mellon University
Pittsburgh, PA 15213
ggordon@cs.cmu.edu

## ABSTRACT

A central problem in artificial intelligence is that of planning to maximize future reward under uncertainty in a partially observable environment. In this paper we propose and demonstrate a novel algorithm which accurately *learns* a model of such an environment directly from sequences of action-observation pairs. We then *close the loop* from observations to actions by planning in the learned model and recovering a policy which is near-optimal in the original environment. Specifically, we present an efficient and statistically consistent spectral algorithm for learning the parameters of a Predictive State Representation (PSR). We demonstrate the algorithm by learning a model of a simulated high-dimensional, vision-based mobile robot planning task, and then perform approximate point-based planning in the learned PSR. Analysis of our results shows that the algorithm learns a state space which efficiently captures the essential features of the environment. This representation allows accurate prediction with a small number of parameters, and enables successful and efficient planning.

## 1. INTRODUCTION

Planning a sequence of actions or a policy to maximize future reward has long been considered a fundamental problem for autonomous agents. For many years, *Partially Observable Markov Decision Processes* (POMDPs) [1, 27, 4] have been considered the most general framework for single agent planning. POMDPs model the state of the world as a latent variable and explicitly reason about uncertainty in both action effects and state observability. Plans in POMDPs are expressed as *policies*, which specify the action to take given any possible probability distribution over state. Unfortunately, exact planning algorithms such as *value iteration* [27] are computationally intractable for most realistic POMDP planning problems. There are arguably two primary reasons for this [18]. The first is the "curse of dimensionality": for a POMDP with $n$ states, the optimal policy is a function of an $n-1$ dimensional distribution over latent state. The second is the "curse of history": the number of distinct policies increases exponentially in the planning horizon. We hope to mitigate the curse of dimensionality by seeking a dynamical system model with *compact dimensionality*, and to mitigate the curse of history by looking for a model that is susceptible to *approximate* planning.

*Predictive State Representations (PSRs)* [13] and the closely related *Observable Operator Models (OOMs)* [9] are generalizations of POMDPs that have attracted interest because they both have greater representational capacity than POMDPs and yield representations that are *at least* as compact [24, 5]. In contrast to the latent-variable representations of POMDPs, PSRs and OOMs represent the state of a dynamical system by tracking occurrence probabilities of a set of future events (called *tests* or *characteristic events*) conditioned on past events (called *histories* or *indicative events*). Because tests and histories are observable quantities, it has been suggested that learning PSRs and OOMs should be easier than learning POMDPs. A final benefit of PSRs and OOMs is that many successful approximate planning techniques for POMDPs can be used to plan in these observable models with minimal adjustment. Accordingly, PSR and OOM models of dynamical systems have potential to overcome both the "curse of dimensionality" (by compactly modeling state), and the "curse of history" (by applying approximate planning techniques).

The quality of an optimized policy for a POMDP, PSR, or OOM depends strongly on the accuracy of the model: inaccurate models typically lead to useless plans. We can specify a model manually or learn one from data, but due to the difficulty of learning, it is far more common to see planning algorithms applied to manually-specified models. Unfortunately, it is usually only possible to hand-specify accurate models for small systems where there is extensive and goal-relevant domain knowledge. For example, recent extensions of approximate planning techniques for PSRs have only been applied to models constructed by hand [11, 8]. For the most part, learning models for planning in partially observable environments has been hampered by the inaccuracy of learning algorithms. For example, Expectation-Maximization (EM) [2] does not avoid local minima or scale to large state spaces; and, although many learning algorithms have been proposed for PSRs [25, 10, 34, 16, 30, 3] and OOMs [9, 6, 14] that attempt to take advantage of the observability of the state representation, none have been shown to learn models that are accurate enough for planning. As a result, there have been few successful attempts at learning a model directly from data and then closing the loop by planning in that model.

Several researchers have, however, made progress in the problem of planning using a learned model. In one instance [21], researchers obtained a POMDP heuristically from the output of a model-free algorithm [15] and demonstrated planning on a small toy maze. In another instance [20], researchers used Markov Chain Monte Carlo (MCMC) inference both to learn a factored Dynamic Bayesian Network (DBN) representation of a POMDP in a small synthetic network administration domain, as well as to perform online planning. Due to the cost of the MCMC sampler used, this approach is still impractical for larger models. In a final example, researchers learned Linear-Linear Exponential Family PSRs from an agent traversing a simulated environment, and found a policy using a policy gradient technique with a parameterized function of the learned PSR staten as input [33, 31]. In this case both the learning and the planning algorithm were subject to local optima. In addition, the authors determined that the learned model was too inaccurate to support value-function-based planning methods [31].

The current paper differs from these and other previous examples of planning in learned models: it both uses a principled and provably statistically consistent model-learning algorithm, and demonstrates positive results on a challenging high-dimensional problem with continuous observations. In particular, we propose a novel, consistent spectral algorithm for learning a variant of PSRs called *Transformed PSRs* [19] directly from execution traces. The algorithm is closely related to subspace identification for learning linear dynamical systems (LDSs) [26, 29] and spectral algorithms for learning Hidden Markov Models (HMMs) [7] and reduced-rank Hidden Markov Models [22]. We then demonstrate that this algorithm is able to learn compact models of a difficult, realistic dynamical system without any prior domain knowledge built into the model or algorithm. Finally, we perform point-based approximate value iteration in the learned compact models, and demonstrate that the greedy policy for the resulting value function works well in the original (not the learned) system. To our knowledge this is the first research that combines all of these achievements, closing the loop from observations to actions in an unknown domain with no human intervention beyond collecting the raw transition data.

## 2. PREDICTIVE STATE REPRESENTATIONS

A predictive state representation (PSR) [13] is a compact and complete description of a dynamical system that represents state as a set of predictions of observable experiments or *tests* that one could perform in the system. Specifically, a test of length $k$ is an ordered sequence of action-observation pairs $\tau = a_1 o_1 \ldots a_k o_k$ that can be executed and observed at a given time. Likewise, a *history* is an ordered sequence of action-observation pairs $h = a_1^h o_1^h \ldots a_t^h o_t^h$ that has been executed and observed prior to a given time. The *prediction* for a test $\tau$ is the probability of the sequence of observations $o_1, \ldots, o_k$ being generated, given that we intervene to take the sequence of actions $a_1, \ldots, a_k$. If the observations produced by the dynamical system match those specified by the test, then the test is said to have *succeeded*. The key idea behind a PSR is that, if the expected outcomes of executing all possible tests are known, then everything there is to know about the state of a dynamical system is also known.

In PSRs, actions in tests are *interventions*, not observations. Thus it is notationally convenient to separate a test $\tau$ into the observation component $\tau^O$ and the action component $\tau^A$. In equations that contain probabilities, a single vertical bar | indicates conditioning and a double vertical bar || indicates intervening. For example, $p(\tau_i^O|h||\tau_i^A)$ is the probability of the observations in test $\tau_i$, conditioned on history $h$, and given that we intervene to execute the actions in $\tau_i$.

Formally a PSR consists of five elements $\{A, O, Q, m_1, F\}$. $A$ is the set of actions that can be executed at each time-step, $O$ is the set of possible observations, and $Q$ is a set of *core tests*. A set of core tests $Q$ has the property that for *any* test $\tau$, there exists some function $f_\tau$ such that $p(\tau^O|h||\tau^A) = f_\tau(p(Q^O|h||Q^A))$ for all histories $h$. Here, the *prediction vector*

$$p(Q^O|h||Q^A) = [p(q_1^O|h||q_1^A), ..., p(q_{|Q|}^O|h||q_{|Q|}^A)]^\mathsf{T} \quad (1)$$

contains the probabilities of success of the tests in $Q$. The existence of $f_\tau$ means that knowing the probabilities for the tests in $Q$ is sufficient for computing the probabilities for all other tests, so the prediction vector is a *sufficient statistic* for the system. The vector $m_1$ is the initial prediction for the outcomes of the tests in $Q$ given some initial distribution over histories $\omega$. We will allow the initial distribution to be general; in practice $\omega$ might correspond to the steady state distribution for a heuristic exploration policy, or the distribution over histories when we first encounter the system, or the empty history with probability 1.

In order to maintain predictions in the tests in $Q$ we need to compute $p(Q^O|ho||a, Q^A)$, the distribution over test outcomes given a new extended history, from the current distribution $p(Q^O|h||Q^A)$ (here $p(Q^O|ho||a, Q^A)$ is the probability over test outcomes conditioned on history $h$ and observation $o$ given the intervention of choosing the immediate next action $a$ and the appropriate actions for the test). Let $f_{aoq}$ be the function needed to update our prediction of test $q \in Q$ given an action $a$ and an observation $o$. (This function is guaranteed to exist since we can set $\tau = aoq$ in $f_\tau$ above.) Finally, $F$ is the set of functions $f_{aoq}$ for all $a \in A$, $o \in O$, and $q \in Q$.

In this work we will restrict ourselves to *linear* PSRs, a subset of PSRs where the functions $f_{aoq}$ are required to be linear in the prediction vector $p(Q^O|h||Q^A)$, so that $f_{aoq}(p(Q^O|h||Q^A)) = m_{aoq}^\mathsf{T} p(Q^O|h||Q^A)$ for some vector $m_{aoq} \in \mathbb{R}^{|Q|}$.[1] We write $M_{ao}$ to be the matrix with rows $m_{aoq}^\mathsf{T}$. By Bayes' Rule, the update from history $h$, after taking action $a$ and seeing observation $o$, is:

$$p(Q^O|ho||a, Q^A) = \frac{p(o, Q^O|h||a, Q^A)}{p(o|h||a)}$$
$$= \frac{M_{ao} p(Q^O|h||Q^A)}{m_\infty^\mathsf{T} M_{ao} p(Q^O|h||Q^A)} \quad (2)$$

where $m_\infty$ is a normalizing vector. Specifying a PSR involves first finding a set of core tests $Q$, called the *discovery problem*, and then finding the parameters $M_{ao}$ and $m_\infty$ for those tests as well as an initial state $m_1$, called the *learning problem*. The discovery problem is usually solved by searching for linearly independent tests by repeatedly performing Singular Value Decompositions (SVDs) on collections of tests [10, 34]. The learning problem is then solved by regression.

---
[1] Linear PSRs have been shown to be a highly expressive class of models [9, 24]: if the set of core tests is *minimal*, then the set of PSRs with $n = |Q|$ core tests is provably *equivalent* to the set of dynamical systems with linear dimension $n$. The *linear dimension* of a dynamical system is a measure of its intrinsic complexity; specifically, it is the rank of the *system-dynamics matrix* [24] of the dynamical system. Since there exist dynamical systems of finite linear dimension which cannot be modeled by any POMDP (or HMM) with a finite number of states (see [9] for an example), POMDPs and HMMs are a proper subset of PSRs [24].

## 2.1 Transformed PSRs

Transformed PSRs (TPSRs) [19] are a generalization of PSRs that maintain a small number of *linear combinations* of test probabilities as sufficient statistics of the dynamical system. As we will see, transformed PSRs can be thought of as *linear transformations* of regular PSRs. Accordingly, TPSRs include PSRs as a special case since this transformation can be the identity matrix. The main benefit of TPSRs is that given a set of core tests, the parameter learning problem can be solved and a large step toward solving the discovery problem can be achieved in closed form. In this respect, TPSRs are closely related to the transformed representations of LDSs and HMMs found by *subspace identification* [29, 26, 7].

For some dynamical system, let $Q$ be the minimal set of core tests with cardinality $n = |Q|$ equal to the dimensionality of the linear system. Then, let $\mathcal{T}$ be a set of core tests (not necessarily minimal) and let $\mathcal{H}$ be a sufficient set of indicative events. A *set of indicative events* is a mutually exclusive and exhaustive partition of the set of all possible histories. We will define a sufficient set of indicative events below. For TPSRs, $|\mathcal{T}|$ and $|\mathcal{H}|$ may be arbitrarily larger than $n$; in practice we might choose $\mathcal{T}$ and $\mathcal{H}$ by selecting sets that we believe to be large enough and varied enough to exhibit the types of behavior that we wish to model.

We define several matrices in terms of $\mathcal{T}$ and $\mathcal{H}$. In each of these matrices we assume that histories $H$ are sampled according to $\omega$; further actions and observations are specified in the individual probability expressions. $P_{\mathcal{H}} \in \mathbb{R}^{|\mathcal{H}|}$ is a vector containing the probabilities of every $h \in \mathcal{H}$.

$$[P_{\mathcal{H}}]_i \equiv \Pr[H \in h_i]$$
$$= \omega(H \in h_i)$$
$$\equiv \pi_{h_i}$$
$$\Rightarrow P_{\mathcal{H}} = \pi \qquad (3a)$$

Here we have defined two notations, $P_{\mathcal{H}}$ and $\pi$, for the same vector. Below we will generalize $P_{\mathcal{H}}$, but keep the same meaning for $\pi$.

Next we define $P_{\mathcal{T},\mathcal{H}} \in \mathbb{R}^{|\mathcal{T}| \times |\mathcal{H}|}$, a matrix with entries that contain the *joint* probability of every test $\tau_i \in \mathcal{T}$ ($1 \le i \le |\mathcal{T}|$) and every indicative event $h_j \in \mathcal{H}$ ($1 \le j \le |\mathcal{H}|$) (assuming we execute test actions $\tau_i^A$):

$$[P_{\mathcal{T},\mathcal{H}}]_{i,j} \equiv \Pr[\tau_i^O, H \in h_j || \tau_i^A]$$
$$= \Pr[\tau_i^O | H \in h_j || \tau_i^A] \Pr[H \in h_j]$$
$$\equiv r_{\tau_i}^\mathsf{T} \Pr[Q^O | H \in h_j || Q^A] \Pr[H \in h_j]$$
$$\equiv r_{\tau_i}^\mathsf{T} s_{h_j} \Pr[H \in h_j]$$
$$= r_{\tau_i}^\mathsf{T} s_{h_j} \pi$$
$$\Rightarrow P_{\mathcal{T},\mathcal{H}} = RS\mathrm{diag}(\pi) \qquad (3b)$$

The vector $r_{\tau_i}$ is the linear function that specifies the probability of the test $\tau_i$ given the probabilities of core tests $Q$. The vector $s_{h_j}$ contains the probabilities of all core tests $Q$ given that the history belongs to the indicative event $h_j$. Because of our assumptions about the linear dimension of the system, the matrix $P_{\mathcal{T},\mathcal{H}}$ factors according to $R \in \mathbb{R}^{|\mathcal{T}| \times n}$ (a matrix with rows $r_{\tau_i}^\mathsf{T}$ for all $1 \le i \le |\mathcal{T}|$) and $S \in \mathbb{R}^{n \times |\mathcal{H}|}$ (a matrix with columns $s_{h_j}$ for all $1 \le j \le |\mathcal{H}|$). Therefore, the *rank* of $P_{\mathcal{T},\mathcal{H}}$ is no more than the linear dimension of the system. At this point we can define a *sufficient* set of indicative events as promised: it is a set of indicative events which ensures that the rank of $P_{\mathcal{T},\mathcal{H}}$ is *equal* to the linear dimension of the system. Finally, $m_1$, which we have defined as the initial prediction for the outcomes of tests in $Q$ given some initial distribution over histories $h$, is given by $m_1 = S\pi$ (here we are taking the expectation of the columns of $S$ according to the correct distribution over histories $\omega$).

We define $P_{\mathcal{T},ao,\mathcal{H}} \in \mathbb{R}^{|\mathcal{T}| \times |\mathcal{H}|}$, a set of matrices, one for each action-observation pair, that represent the probabilities of a *triple* of an indicative event $h_j$, the immediate following observation $O$, and a subsequent test $\tau_j$, given the appropriate actions:

$$[P_{\mathcal{T},ao,\mathcal{H}}]_{i,j} \equiv \Pr[\tau_i^O, O = o, H \in h_j || A = a, \tau_i^A]$$
$$= \Pr[\tau_i^O, O = o | H \in h_j || A = a, \tau_i^A] \Pr[H \in h_j]$$
$$= \Pr[\tau_i^O | H \in h_j, O = o || A = a, \tau_i^A]$$
$$\quad \Pr[O = o | H \in h_j || A = a] \Pr[H \in h_j]$$
$$= r_{\tau_i}^\mathsf{T} \Pr[Q^O | H \in h_j, O = o || A = a, Q^A]$$
$$\quad \Pr[O = o | H \in h_j || A = a] \Pr[H \in h_j]$$
$$= r_{\tau_i}^\mathsf{T} M_{ao} \Pr[Q^O | H \in h_j || Q^A] \Pr[H \in h_j]$$
$$= r_{\tau_i}^\mathsf{T} M_{ao} s_{h_j} \Pr[H \in h_j]$$
$$= r_{\tau_i}^\mathsf{T} M_{ao} s_{h_j} \pi_{h_j}$$
$$\Rightarrow P_{\mathcal{T},ao,\mathcal{H}} = RM_{ao}S\mathrm{diag}(\pi) \qquad (3c)$$

The matrices $P_{\mathcal{T},ao,\mathcal{H}}$ factor according to R and S (defined above) and the PSR transition matrix $M_{ao} \in \mathbb{R}^{n \times n}$. Note that $R$ spans the column space of both $P_{\mathcal{T},\mathcal{H}}$ and the matrices $P_{\mathcal{T},ao,\mathcal{H}}$; we make use of this fact below.

Finally, we will use the fact that $m_\infty$ is a normalizing vector to derive the equations below (by repeatedly multiplying by $S$ and $S^\dagger$, and using the facts $SS^\dagger = I$ and $m_\infty^\mathsf{T} S = 1^\mathsf{T}$, since each column of $S$ is a vector of core-test predictions). Here, $k = |\mathcal{H}|$ and $1_k$ denotes the ones-vector of length $k$:

$$m_\infty^\mathsf{T} S = 1_k^\mathsf{T}$$
$$m_\infty^\mathsf{T} SS^\dagger = 1_k^\mathsf{T} S^\dagger$$
$$m_\infty^\mathsf{T} = 1_k^\mathsf{T} S^\dagger \qquad (4a)$$
$$m_\infty^\mathsf{T} S = 1_k^\mathsf{T} S^\dagger S$$
$$1_k^\mathsf{T} = 1_k^\mathsf{T} S^\dagger S \qquad (4b)$$

We now define a TPSR in terms of the matrices $P_{\mathcal{H}}$, $P_{\mathcal{T},\mathcal{H}}$, $P_{\mathcal{T},ao,\mathcal{H}}$ and an additional matrix $U$ that obeys the condition that $U^\mathsf{T} R$ is invertible. In other words, the columns of $U$ define an $n$-dimensional subspace that is *not* orthogonal to the column space of $P_{\mathcal{T},\mathcal{H}}$. A natural choice for $U$ is given by the left singular vectors of $P_{\mathcal{T},\mathcal{H}}$.

With these definitions, we define the parameters of a TPSR in terms of observable matrices and simplify the expressions using Equations 3(a–c), as follows (here, $B_{ao}$ is a similarity transform of the low-dimensional linear transition matrix $M_{ao}$ and $b_1$ and $b_\infty$ are the corresponding linear transformations of the minimal PSR initial state $M_1$ and the normalizing vector):

$$b_1 \equiv U^\mathsf{T} P_{\mathcal{T},\mathcal{H}} 1_k$$
$$= U^\mathsf{T} RS\mathrm{diag}(\pi) 1_k$$
$$= U^\mathsf{T} RS\pi$$
$$= (U^\mathsf{T} R) m_1 \qquad (5a)$$

$$\begin{aligned}
b_\infty^\mathsf{T} &\equiv P_\mathcal{H}^\mathsf{T}(U^\mathsf{T} P_{\mathcal{T},\mathcal{H}})^\dagger \\
&= 1_n^\mathsf{T} S^\dagger S \mathrm{diag}(\pi)(U^\mathsf{T} P_{\mathcal{T},\mathcal{H}})^\dagger \\
&= 1_n^\mathsf{T} S^\dagger (U^\mathsf{T} R)^{-1}(U^\mathsf{T} R) S \mathrm{diag}(\pi)(U^\mathsf{T} P_{\mathcal{T},\mathcal{H}})^\dagger \\
&= 1_n^\mathsf{T} S^\dagger (U^\mathsf{T} R)^{-1} U^\mathsf{T} P_{\mathcal{T},\mathcal{H}} (U^\mathsf{T} P_{\mathcal{T},\mathcal{H}})^\dagger \\
&= 1_n^\mathsf{T} S^\dagger (U^\mathsf{T} R)^{-1} \\
&= m_\infty^\mathsf{T} (U^\mathsf{T} R)^{-1} \quad (5b)\\
B_{ao} &\equiv U^\mathsf{T} P_{\mathcal{T},ao,\mathcal{H}} (U^\mathsf{T} P_{\mathcal{T},\mathcal{H}})^\dagger \\
&= U^\mathsf{T} R M_{ao} S \mathrm{diag}(\pi)(U^\mathsf{T} P_{\mathcal{T},\mathcal{H}})^\dagger \\
&= U^\mathsf{T} R M_{ao} (U^\mathsf{T} R)^{-1}(U^\mathsf{T} R) S \mathrm{diag}(\pi)(U^\mathsf{T} P_{\mathcal{T},\mathcal{H}})^\dagger \\
&= (U^\mathsf{T} R) M_{ao} (U^\mathsf{T} R)^{-1} U^\mathsf{T} P_{\mathcal{T},\mathcal{H}} (U^\mathsf{T} P_{\mathcal{T},\mathcal{H}})^\dagger \\
&= (U^\mathsf{T} R) M_{ao} (U^\mathsf{T} R)^{-1} \quad (5c)
\end{aligned}$$

The derivation of Equation 5b makes use of Equations 4a and 4b. Given these parameters we can calculate the probability of observations $o_{1:t}$ at any time $t$ given that we intervened with actions $a_{1:t}$, from the initial state $m_1$. Here we write the product of each $M_{ao}$ (one for each action observation pair) $M_{a_1 o_1} M_{a_2 o_2} \ldots M_{a_t o_t}$ as $M_{ao_{1:t}}$.

$$\begin{aligned}
\Pr[o_{1:t} || a_{1:t}] &= m_\infty^\mathsf{T} M_{ao_{1:t}} m_1 \\
&= m_\infty^\mathsf{T} (U^\mathsf{T} R)^{-1}(U^\mathsf{T} R) M_{ao_{1:t}} (U^\mathsf{T} R)^{-1}(U^\mathsf{T} R) m_1 \\
&= b_\infty^\mathsf{T} B_{ao_{1:t}} b_1 \quad (6)
\end{aligned}$$

In addition to the initial TPSR state $b_1$, we define normalized conditional 'internal states' $b_t$. We define the TPSR state at time $t+1$ as:

$$b_{t+1} \equiv \frac{B_{ao_{1:t}} b_1}{b_\infty^\mathsf{T} B_{ao_{1:t}} b_1} \quad (7)$$

We can define a *recursive* state update for $t > 1$ as follows (using Equation 7 as the base case for $t = 1$):

$$\begin{aligned}
b_{t+1} &\equiv \frac{B_{ao_{1:t}} b_1}{b_\infty^\mathsf{T} B_{ao_{1:t}} b_1} \\
&= \frac{B_{ao_t} B_{ao_{1:t-1}} b_1}{b_\infty^\mathsf{T} B_{ao_t} B_{ao_{1:t-1}} b_1} \\
&= \frac{B_{ao_t} b_t}{b_\infty^\mathsf{T} B_{ao_t} b_t} \quad (8)
\end{aligned}$$

The prediction of tests $p(\mathcal{T}^O | h || \mathcal{T}^A)$ at time $t$ is given by $U b_t = U U^\mathsf{T} R s_t = R s_t$, and the rotation from a TPSR to a PSR is given by $s_t = (U^\mathsf{T} R)^{-1} b_t$ where $s_t$ is the prediction vector for the PSR. Note that in general, the elements of the linear combinations $b_t$ cannot be interpreted as probabilities since they may lie outside the range $[0, 1]$.

## 3. LEARNING TPSRS

Our learning algorithm works by building empirical estimates $\widehat{P}_\mathcal{H}$, $\widehat{P}_{\mathcal{T},\mathcal{H}}$, and $\widehat{P}_{\mathcal{T},ao,\mathcal{H}}$ of the matrices $P_\mathcal{H}$, $P_{\mathcal{T},\mathcal{H}}$, and $P_{\mathcal{T},ao,\mathcal{H}}$ defined above. To build these estimates, we repeatedly sample a history $h$ from the distribution $\omega$, execute a sequence of actions, and record the resulting observations. This data gathering strategy implies that we must be able to arrange for the system to be in a state corresponding to $h \sim \omega$; for example, if our system has a reset, we can take $\omega$ to be the distribution resulting from executing a fixed exploration policy for a few steps after reset.

In practice, reset is often not available. In this case we can estimate $\widehat{P}_\mathcal{H}$, $\widehat{P}_{\mathcal{T},\mathcal{H}}$, and $\widehat{P}_{\mathcal{T},ao,\mathcal{H}}$ by dividing a single long sequence of action-observation pairs into subsequences and pretending that each subsequence started with a reset. We are forced to use an initial distribution over histories, $\omega$, equal to the steady state distribution of the policy which generated the data. This approach is called the *suffix-history* algorithm [34]. With this method, the estimated matrices will be only approximately correct, since interventions that we take at one time will affect the distribution over histories at future times; however, the approximation is often a good one in practice.

Once we have computed $\widehat{P}_\mathcal{H}$, $\widehat{P}_{\mathcal{T},\mathcal{H}}$, and $\widehat{P}_{\mathcal{T},ao,\mathcal{H}}$, we can generate $\widehat{U}$ by singular value decomposition of $\widehat{P}_{\mathcal{T},\mathcal{H}}$. We can then learn the TPSR parameters by plugging $\widehat{U}$, $\widehat{P}_\mathcal{H}$, $\widehat{P}_{\mathcal{T},\mathcal{H}}$, and $\widehat{P}_{\mathcal{T},ao,\mathcal{H}}$ into Equation 5. For reference, we summarize the above steps here[2]:

1. Compute empirical estimates $\widehat{P}_\mathcal{H}$, $\widehat{P}_{\mathcal{T},\mathcal{H}}$, $\widehat{P}_{\mathcal{T},ao,\mathcal{H}}$.

2. Use SVD on $\widehat{P}_{\mathcal{T},\mathcal{H}}$ to compute $\widehat{U}$, the matrix of left singular vectors corresponding to the $n$ largest singular values.

3. Compute model parameter estimates:

   (a) $\widehat{b}_1 = \widehat{U}^\mathsf{T} \widehat{P}_\mathcal{H}$,

   (b) $\widehat{b}_\infty = (\widehat{P}_{\mathcal{T},\mathcal{H}}^\mathsf{T} \widehat{U})^\dagger \widehat{P}_\mathcal{H}$,

   (c) $\widehat{B}_{ao} = \widehat{U}^\mathsf{T} \widehat{P}_{\mathcal{T},ao,\mathcal{H}} (\widehat{U}^\mathsf{T} \widehat{P}_{\mathcal{T},\mathcal{H}})^\dagger$

As we include more data in our averages, the law of large numbers guarantees that our estimates $\widehat{P}_\mathcal{H}$, $\widehat{P}_{\mathcal{T},\mathcal{H}}$, and $\widehat{P}_{\mathcal{T},ao,\mathcal{H}}$ converge to the true matrices $P_\mathcal{H}$, $P_{\mathcal{T},\mathcal{H}}$, and $P_{\mathcal{T},ao,\mathcal{H}}$ (defined in Equation 3). So by continuity of the formulas in steps 3(a–c) above, if our system is truly a TPSR of finite rank, our estimates $\widehat{b}_1$, $\widehat{b}_\infty$, and $\widehat{B}_{ao}$ converge to the true parameters up to a linear transform. Although parameters estimated with finite data can sometimes lead to negative probability estimates when filtering or predicting, this can be avoided in practice by thresholding the prediction vectors by some small positive probability.

Note that the learning algorithm presented here is distinct from the TPSR learning algorithm presented in Rosencrantz et al. [19]. The principal difference between the two algorithms is that here we estimate the *joint* probability of a past event, a current observation, and a future event in the matrix $\widehat{P}_{\mathcal{T},ao,\mathcal{H}}$ whereas in [19], the authors instead estimate the probability of a future event, *conditioned* on a past event and a current observation. To compensate, Rosencrantz et al. later multiply this estimate by an approximation of the probability of the current observation, conditioned on the past event, but not until after the SVD is applied. Rosencrantz et al. also derive the approximate probability of the current observation differently: as the result of a regression instead of directly from empirical counts. Finally, Rosencrantz et al. do not make any attempt to multiply by the marginal probability of the past event, although this term cancels in the current work so it is possible that, in the absence of estimation errors, both algorithms arrive at the same answer.

---

[2]The learning strategy employed here may be seen as a generalization of Hsu et al.'s spectral algorithm for learning HMMs [7] to PSRs. Note that since HMMs and POMDPs are a proper subset of PSRs, we can use the algorithm in this paper to learn back both HMMs and POMDPs in PSR form.

Below we present two extensions to our learning algorithm that preserve consistency while relaxing the requirement that we find a discrete set of indicative events and tests. These extensions make learning substantially easier for many difficult domains (e.g. for continuous observations) in practice.

### 3.1 Learning TPSRs with Indicative and Characteristic Features

In data gathered from complex real-world dynamical systems, it may not be possible to find a reasonably-sized set of discrete core tests $\mathcal{T}$ or indicative events $\mathcal{H}$. When this is the case, we can generalize the TPSR learning algorithm and work with *features* of tests and histories, which we call *characteristic features* and *indicative features* respectively. In particular let $\mathcal{T}$ and $\mathcal{H}$ be large sets of tests and indicative events (possibly too large to work with directly) and let $\phi^\mathcal{T}$ and $\phi^\mathcal{H}$ be shorter vectors of characteristic and indicative features. The matrices $P_\mathcal{H}$, $P_{\mathcal{T},\mathcal{H}}$, and $P_{\mathcal{T},ao,\mathcal{H}}$ will no longer contain probabilities but rather *expected values* of features or products of features. For the special case of features that are *indicator functions* of tests and histories, we recover the TPSR matrices from Section 2.1 where $P_\mathcal{H}$, $P_{\mathcal{T},\mathcal{H}}$, and $P_{\mathcal{T},ao,\mathcal{H}}$ consist of probabilities.

Here we prove the *consistency* of our estimation algorithm using these more general matrices as inputs. In the following equations $\Phi^\mathcal{T}$ and $\Phi^\mathcal{H}$ are matrices of characteristic and indicative features respectively, with first dimension equal to the number of characteristic or indicative features and second dimension equal to $|\mathcal{T}|$ and $|\mathcal{H}|$ respectively.

An entry of $\Phi^\mathcal{H}$ is the expectation of one of the indicative features given the occurrence of one of the indicative events. An entry of $\Phi^\mathcal{T}$ is the weight of one of our tests in calculating one of our characteristic features. With these features we generalize the matrices $P_\mathcal{H}$, $P_{\mathcal{T},\mathcal{H}}$, and $P_{\mathcal{T},ao,\mathcal{H}}$:

$$[P_\mathcal{H}]_i \equiv \mathbb{E}(\phi^\mathcal{H}_i(h)) = \sum_{h \in \mathcal{H}} \Pr[H \in h]\Phi^\mathcal{H}_{ih}$$

$$\Rightarrow P_\mathcal{H} = \Phi^\mathcal{H} \pi \quad (9a)$$

$$[P_{\mathcal{T},\mathcal{H}}]_{i,j} \equiv \mathbb{E}(\phi^\mathcal{T}_i(\tau^O) \cdot \phi^\mathcal{H}_j(h)||\tau^A)$$

$$= \sum_{\tau \in \mathcal{T}} \sum_{h \in \mathcal{H}} \Pr[\tau^O, H \in h||\tau^A]\Phi^\mathcal{T}_{i\tau}\Phi^\mathcal{H}_{jh}$$

$$= \sum_{\tau \in \mathcal{T}} \sum_{h \in \mathcal{H}} r^\mathsf{T}_\tau s_h \pi_h \Phi^\mathcal{T}_{i\tau}\Phi^\mathcal{H}_{jh} \quad \text{(by Eq. (3b))}$$

$$= \sum_{\tau \in \mathcal{T}} r^\mathsf{T}_\tau \Phi^\mathcal{T}_{i\tau} \sum_{h \in \mathcal{H}} s_h \pi_h \Phi^\mathcal{H}_{jh}$$

$$\Rightarrow P_{\mathcal{T},\mathcal{H}} = \Phi^\mathcal{T} RS \mathrm{diag}(\pi)\Phi^{\mathcal{H}\mathsf{T}} \quad (9b)$$

$$[P_{\mathcal{T},ao,\mathcal{H}}]_{i,j} \equiv \mathbb{E}(\phi^\mathcal{T}_i(\tau^O) \cdot \phi^\mathcal{H}_j(h) \cdot \delta(O=o)||\tau^A A = a)$$

$$= \sum_{\tau \in \mathcal{T}} \sum_{h \in \mathcal{H}} \Pr[\tau^O, O=o, H \in h||A=a, \tau^A]\Phi^\mathcal{T}_{i\tau}\Phi^\mathcal{H}_{jh}$$

$$= \sum_{\tau \in \mathcal{T}} \sum_{h \in \mathcal{H}} r^\mathsf{T}_\tau M_{ao} s_h \pi_h \Phi^\mathcal{T}_{i\tau}\Phi^\mathcal{H}_{jh} \quad \text{(by Eq. (3c))}$$

$$= \left(\sum_{\tau \in \mathcal{T}} r^\mathsf{T}_\tau \Phi^\mathcal{T}_{i\tau}\right) M_{ao} \left(\sum_{h \in \mathcal{H}} s_h \pi_h \Phi^\mathcal{H}_{jh}\right)$$

$$\Rightarrow P_{\mathcal{T},ao,\mathcal{H}} = \Phi^\mathcal{T} R M_{ao} S \mathrm{diag}(\pi)\Phi^{\mathcal{H}\mathsf{T}} \quad (9c)$$

where $\delta(O=o)$ is an indicator function for a particular observation. The parameters of the TPSR are defined in terms of a matrix $U$ that obeys the condition that $U^\mathsf{T}\Phi^\mathcal{T} R$ is invertible (we can take $U$ to be the left singular values of $P_{\mathcal{T},\mathcal{H}}$), and in terms of the matrices $P_\mathcal{H}$, $P_{\mathcal{T},\mathcal{H}}$, and $P_{\mathcal{T},ao,\mathcal{H}}$. We also define a new vector $e$ s.t. $\Phi^{\mathcal{H}\mathsf{T}} e^\mathsf{T} = 1_k$; this means that the ones vector $1_k^\mathsf{T}$ must be in the row space of $\Phi^\mathcal{H}$. Since $\Phi^\mathcal{H}$ is a matrix of features, we can always ensure that this is the case by requiring one of our features to be a constant. Then, one row of $\Phi^\mathcal{H}$ is $1_k^\mathsf{T}$, and we can set $e^\mathsf{T} = \begin{bmatrix} 1 & 0 & \ldots & 0 \end{bmatrix}^\mathsf{T}$. Finally we define the generalized TPSR parameters $b_1$, $b_\infty$, and $B_{ao}$ as follows:

$$b_1 \equiv U^\mathsf{T} P_{\mathcal{T},\mathcal{H}} e^\mathsf{T}$$

$$= U^\mathsf{T}\Phi^\mathcal{T} RS\mathrm{diag}(\pi)\Phi^{\mathcal{H}\mathsf{T}} e^\mathsf{T}$$

$$= U^\mathsf{T}\Phi^\mathcal{T} RS\mathrm{diag}(\pi)1_k$$

$$= (U^\mathsf{T}\Phi^\mathcal{T} R) S\pi$$

$$= (U^\mathsf{T}\Phi^\mathcal{T} R) m_1 \quad (10a)$$

$$b_\infty^\mathsf{T} \equiv P_\mathcal{H}^\mathsf{T}(U^\mathsf{T} P_{\mathcal{T},\mathcal{H}})^\dagger$$

$$= 1_n^\mathsf{T}\mathrm{diag}(\pi)\Phi^{\mathcal{H}\mathsf{T}}(U^\mathsf{T} P_{\mathcal{T},\mathcal{H}})^\dagger$$

$$= 1_n^\mathsf{T} S^\dagger S\mathrm{diag}(\pi)\Phi^{\mathcal{H}\mathsf{T}}(U^\mathsf{T} P_{\mathcal{T},\mathcal{H}})^\dagger$$

$$= 1_n^\mathsf{T} S^\dagger (U^\mathsf{T}\Phi^\mathcal{T} R)^{-1}(U^\mathsf{T}\Phi^\mathcal{T} R)S\mathrm{diag}(\pi)\Phi^{\mathcal{H}\mathsf{T}}(U^\mathsf{T} P_{\mathcal{T},\mathcal{H}})^\dagger$$

$$= 1_n^\mathsf{T} S^\dagger (U^\mathsf{T}\Phi^\mathcal{T} R)^{-1} U^\mathsf{T} P_{\mathcal{T},\mathcal{H}}(U^\mathsf{T} P_{\mathcal{T},\mathcal{H}})^\dagger$$

$$= 1_n^\mathsf{T} S^\dagger (U^\mathsf{T}\Phi^\mathcal{T} R)^{-1}$$

$$= m_\infty^\mathsf{T}(U^\mathsf{T}\Phi^\mathcal{T} R)^{-1} \quad (10b)$$

$$B_{ao} \equiv U^\mathsf{T} P_{\mathcal{T},ao,\mathcal{H}}(U^\mathsf{T} P_{\mathcal{T},\mathcal{H}})^\dagger$$

$$= U^\mathsf{T}\Phi^\mathcal{T} RM_{ao}S\mathrm{diag}(\pi)\Phi^{\mathcal{H}\mathsf{T}}(U^\mathsf{T} P_{\mathcal{T},\mathcal{H}})^\dagger$$

$$= U^\mathsf{T}\Phi^\mathcal{T} RM_{ao}(U^\mathsf{T}\Phi^\mathcal{T} R)^{-1}(U^\mathsf{T}\Phi^\mathcal{T} R)S\mathrm{diag}(\pi)\Phi^{\mathcal{H}\mathsf{T}}(U^\mathsf{T} P_{\mathcal{T},\mathcal{H}})^\dagger$$

$$= (U^\mathsf{T}\Phi^\mathcal{T} R)M_{ao}(U^\mathsf{T}\Phi^\mathcal{T} R)^{-1} U^\mathsf{T} P_{\mathcal{T},\mathcal{H}}(U^\mathsf{T} P_{\mathcal{T},\mathcal{H}})^\dagger$$

$$= (U^\mathsf{T}\Phi^\mathcal{T} R)M_{ao}(U^\mathsf{T}\Phi^\mathcal{T} R)^{-1} \quad (10c)$$

Just as in the beginning of Section 3, we can estimate $\widehat{P}_\mathcal{H}$, $\widehat{P}_{\mathcal{T},\mathcal{H}}$, and $\widehat{P}_{\mathcal{T},ao,\mathcal{H}}$, and then plug the matrices into Equations 10(a–c). Thus we see that if we work with characteristic and indicative features, and if our system is truly a TPSR of finite rank, our estimates $\widehat{b}_1$, $\widehat{b}_\infty$, and $\widehat{B}_{ao}$ again converge to the true PSR parameters up to a linear transform.

### 3.2 Kernel Density Estimation for Continuous Observations

For continuous observations, we use Kernel Density Estimation (KDE) [23] to model the observation probability density function (PDF). We use a fraction of the training data points as kernel centers, placing one multivariate Gaussian kernel at each point.[3] The KDE estimator of the observation PDF is a convex combination of these kernels; since each kernel integrates to 1, this estimator also integrates to 1. KDE theory [23] tells us that, with the correct kernel weights, as the number of kernel centers and the number of samples go to infinity and the kernel bandwidth goes to zero (at appropriate rates), the KDE estimator converges to the observation PDF in $L_1$ norm. The kernel density estimator is completely determined by the normalized vector of kernel weights; therefore, if we can estimate this vector accurately, our estimate of the observation PDF will converge to the observation PDF as well. Hence our goal is to predict the correct expected value of this normalized kernel vector given all past observations. In the continuous-observation case, we can still write our latent-state update in the same form, using a matrix $B_{ao}$; however, rather than learning each of the uncountably-many $B_{ao}$ matrices separately, we learn one base operator per kernel center, and use convex combinations of these base operators to compute observable operators as needed. For more details on practical aspects of the learning procedure with continuous observations, see Section 5.2.

---

[3]We use a general elliptical covariance matrix, chosen by PCA: that is, we use a spherical covariance after projecting onto the eigenvectors of the covariance matrix of the observations, and scaling by the square roots of the eigenvalues.

## 4. PLANNING IN TPSRS

The primary motivation for modeling a controlled dynamical system is for reasoning about the effects of taking a sequence of actions in the system. The TPSR model can be augmented for this purpose by specifying a reward function for taking an action $a$ in state $b$:

$$\mathcal{R}(b, a) = \eta_a^\mathsf{T} b \tag{11}$$

where $\eta_a^\mathsf{T} \in \mathbb{R}^n$ is the linear reward function for taking action $a$. Given this function and a discount factor $\gamma$, the planning problem for TPSRs is to find a policy that maximizes the expected discounted sum of rewards $\mathbb{E}\left[\sum_t \gamma^t \mathcal{R}(b_t, a_t)\right]$. The optimal policy can be compactly represented using the optimal value function $V^*$, which is defined recursively as:

$$V^*(b) = \max_{a \in A} \left[\mathcal{R}(b, a) + \gamma \sum_{o \in O} p(o|b, a) V^*(b_{ao})\right] \tag{12}$$

where $b_{ao}$ is the state obtained from $b$ after executing action $a$ and observing $o$. When optimized exactly, this value function is always piecewise linear and convex (PWLC) in the state and has finitely many pieces in finite-horizon planning problems.[4] The optimal action is then obtained by taking the arg max instead of the max in Equation 12.

Exact value iteration in POMDPs or TPSRs optimizes the value function over all possible belief or state vectors. Computing the exact value function is problematic because the number of sequences of actions that must be considered grows exponentially with the planning horizon, called the "curse of history." Approximate point-based planning techniques (see below) attempt only to calculate the best sequence of actions at some finite set of belief points. Unfortunately, in high dimensions, approximate planning techniques have difficulty adequately sampling the space of possible beliefs. This is due to the "curse of dimensionality." Because TPSRs often admit a compact low-dimensional representation, approximate point-based planning techniques can work well in these models.

*Point-Based Value Iteration (PBVI)* [17] is an efficient approximation of exact value iteration that performs value backup steps on a finite set of heuristically-chosen belief points rather than over the entire belief simplex. PBVI exploits the fact that the value function is PWLC. A linear lower bound on the value function at one point $b$ can be used as a lower bound at nearby points; this insight allows the value function to be approximated with a finite set of hyperplanes (often called $\alpha$-vectors), one for each point. Although PBVI was designed for POMDPs, the approach has been generalized to PSRs [8]. Formally, given some set of points $\mathcal{B} = \{b_0, \ldots, b_k\}$ in the TPSR state space, we recursively compute the value function and linear lower bounds at only these points. The approximation of the value function can be represented by a set $\Gamma = \{\alpha_0, \ldots, \alpha_k\}$ such that each $\alpha_i$ corresponds to the optimal value function at at least one prediction vector $b_i$. To obtain the approximate value function $V_{t+1}(b)$ from the previous value function $V_t(b)$ we apply the *recursive backup operator* on points in $\mathcal{B}$: if $V_t(b) = \max_{\alpha \in \Gamma_t} \alpha^\mathsf{T} b$, then

$$V_{t+1}(b) = \max_{a \in A} \left[\mathcal{R}(b, a) + \gamma \sum_{o \in O} \max_{\alpha \in \Gamma_t} \alpha^\mathsf{T} B_{ao} b\right] \tag{13}$$

In addition to being tractable on much larger-scale planning problems than exact value iteration, PBVI comes with theoretical guarantees in the form of error bounds that are low-order polynomials in the degree of approximation, range of reward values, and discount factor $\gamma$ [17, 8]. Perseus [28, 11] is a variant of PBVI that updates the value function over a small randomized subset of a large set of reachable belief points at each time step. By only updating a subset of belief points, Perseus can achieve a computational advantage over plain PBVI in some domains. We use Perseus in this paper due to its speed and simplicity of implementation.

## 5. EXPERIMENTAL RESULTS

We have introduced a novel algorithm for learning TPSRs directly from data, as well as a kernel-based extension for modeling continuous observations, and discussed how to plan in the learned model. First we demonstrate the *viability* of this approach to planning in a challenging non-linear, partially observable, controlled domain by learning a model directly from sensor inputs and then "closing the loop" by planning in the learned model. Second, unlike previous attempts to learn PSRs, which either lack planning results [19, 32], or which compare policies within the learned system [33], we compare our resulting policy to a bound on the best possible solution in the original system and demonstrate that the policy is close to optimal.

### 5.1 The Autonomous Robot Domain

The simulated autonomous robot domain consists of a simple $45 \times 45$ unit square arena with a central obstacle and brightly colored walls (Figure 1(A-B)). We modeled the robot as a sphere of radius 2 units. The robot can move around the floor of the arena, and rotate to face in any direction. The robot has a simulated $16 \times 16$ pixel color camera, whose focal plane is located one unit in front of the robot's center of rotation. The robot's visual field was $45°$ in both azimuth and elevation, thus providing the robot with an angular resolution of $\sim 2.8°$ per pixel. Images on the sensor matrix at any moment were simulated by a non-linear perspective transformation of the projected values arising from the robot's position and orientation in the environment at that time. The resulting 768-element pattern of unprocessed RGB values was the only input to an robot (images were *not* preprocessed to extract features), and each action produced a new set of pixel values. The robot was able to move forward 1 or 0 units, and simultaneously rotate 15°, −15°, or 0°, resulting in 6 unique actions. In the real world, friction, uneven surfaces, and other factors confound precisely predictable movements. To simulate this uncertainty, a small amount of Gaussian noise was added to the translation and rotation components of the actions. The robot was allowed to occupy any real-valued $(x, y, \theta)$ pose in the environment, but was not allowed to intersect walls. In case of a collision, we interrupted the current motion just before the robot intersected an obstacle, simulating an inelastic collision.

---
[4]This observation follows from that fact that a TPSR is a linear transformation of a PSR, and PSRs like POMDPs have PWLC value functions [11].

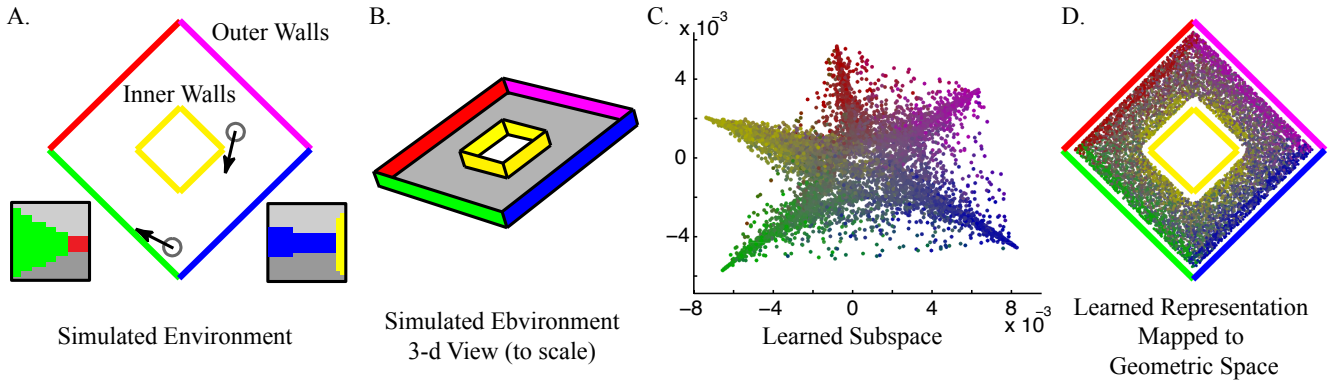

Figure 1: Learning the Autonomous Robot Domain. (A) The robot uses visual sensing to traverse a square domain with multi-colored walls and a central obstacle. Examples of images recorded by the robot occupying two different positions in the environment are shown on the at the bottom of the figure. (B) A to-scale 3-dimensional view of the environment. (C) The 2nd and 3rd dimension of the learned subspace (the first dimension primarily contained normalization information). Each point is the embedding of a single history, displayed with color equal to the average RGB color in the first image in the highest probability test. (D) The same points in (C) projected onto the environment's geometric space.

## 5.2 Learning a Model

We learn our model from a sample of 10000 short trajectories, each containing 7 action-observation pairs. We generate each trajectory by starting from a uniformly randomly sampled position in the environment and executing a uniform random sequence of actions. We used the first $l = 2000$ trajectories to generate kernel centers, and the remaining $w = 8000$ to estimate the matrices $P_\mathcal{H}$, $P_{\mathcal{T},\mathcal{H}}$, and $P_{\mathcal{T},ao,\mathcal{H}}$.

To define these matrices, we need to specify a set of indicative features, a set of observation kernel centers, and a set of characteristic features. We use Gaussian kernels to define our indicative and characteristic features, in a similar manner to the Gaussian kernels described above for observations; our analysis allows us to use arbitrary indicative and characteristic features, but we found Gaussian kernels to be convenient and effective. Note that the resulting features over tests and histories are just *features*; unlike the kernel centers defined over observations, there is no need to let the kernel width approach zero, since we are not attempting to learn accurate PDFs over the histories and tests in $\mathcal{H}$ and $\mathcal{T}$.

In more detail, we define a set of 2000 *indicative kernels*, each one centered at a sequence of 3 observations from the initial segment of one of our trajectories. We choose the kernel covariance using PCA on these sequences of observations, just as described for single observations in Section 3.2. We then generate our indicative features for a new sequence of three observations by evaluating each indicative kernel at the new sequence, and normalizing so that the vector of features sums to one. Similarly, we define 2000 *characteristic kernels*, each one centered at a sequence of 3 observations from the end of one of our sample trajectories, choose a kernel covariance, and define our characteristic feature vector by evaluating each kernel at a new observation sequence and normalizing. The initial distribution $\omega$ is, therefore, the distribution obtained by initializing uniformly and taking 3 random actions. Finally, we define 500 *observation kernels*, each one centered at a single observation from the middle of one of our sample trajectories, and replace each observation by its corresponding vector of normalized kernel weights.

Next, we construct the matrices $\widehat{P}_\mathcal{H}$, $\widehat{P}_{\mathcal{T},\mathcal{H}}$, and $\widehat{P}_{\mathcal{T},ao,\mathcal{H}}$. As defined above, each element of $\widehat{P}_\mathcal{H}$ is the empirical expectation (over our 8,000 training trajectories) of the corresponding element of the indicative feature vector—that is, element $i$ is $\frac{1}{w}\sum_{t=1}^{w} \phi_{it}^\mathcal{H}$, where $\phi_{it}^\mathcal{H}$ is the $i$th indicative feature, evaluated at the current history at time $t$. Similarly, each element of $\widehat{P}_{\mathcal{T},\mathcal{H}}$ is the empirical expectation of the *product* of one indicative feature and one characteristic feature: element $i, j$ is $\frac{1}{w}\sum_{t=1}^{w} \phi_{it}^\mathcal{T} \phi_{jt}^\mathcal{H}$. Once we have constructed $\widehat{P}_{\mathcal{T},\mathcal{H}}$, we can compute $\widehat{U}$ as the matrix of left singular vectors of $\widehat{P}_{\mathcal{T},\mathcal{H}}$. One of the advantages of subspace identification is that the complexity of the model can be tuned by selecting the number of singular vectors in $\widehat{U}$. To learn an *exact* TPSR, we should pick the first $n$ singular vectors that correspond to singular values in $\widehat{P}_{\mathcal{T},\mathcal{H}}$ greater than some cutoff that varies with the noise resolution of our data. However, we may wish to pick a smaller set of singular vectors; doing so will produce a more compact TPSR at the possible loss of prediction quality. We chose $n = 5$, the smallest TPSR that was able to produce high quality policies (see Section 5.4 below).

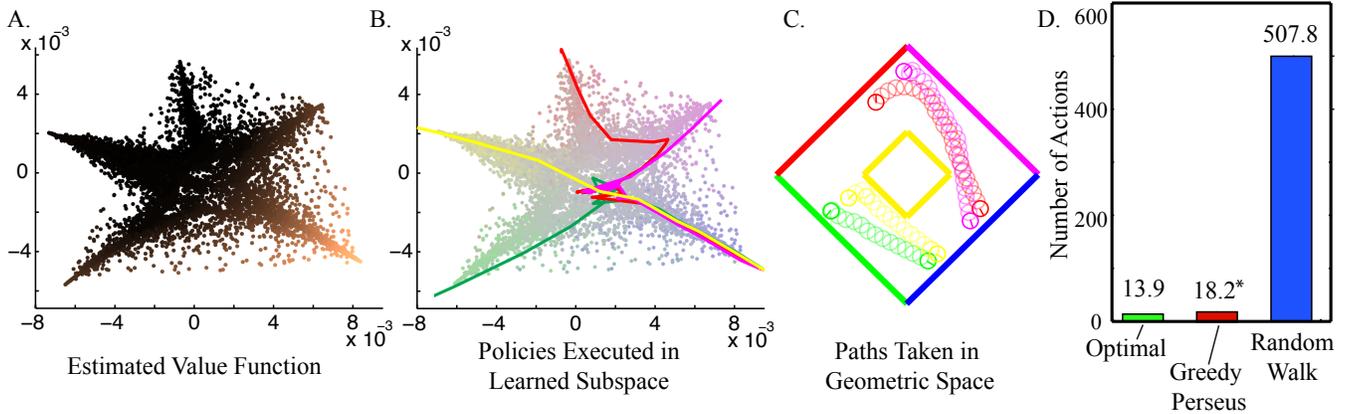

Figure 2: Planning in the Learned State Space. (A) The value function computed for each embedded point; lighter indicates higher value. (B) Policies executed in the learned subspace. The red, green, magenta, and yellow paths correspond to the policy executed by a robot with starting positions facing the red, green, magenta, and yellow walls respectively. (C) The paths taken by the robot in geometric space while executing the policy. Each of the paths corresponds to the path of the same color in (B). The darker circles indicate the starting and ending position of each path, and the tick-mark in the circles indicates the robot's orientation. (D) Mean number of actions in path from 100 randomly sampled start position to the target image (facing blue wall). The first bar (left) is the mean number of actions in the optimal solution found by A* search in the robot's configuration space. The second bar (center) is the mean number of actions taken by executing the policy computed by Perseus in the learned model (the asterisk indicates that this mean was only computed over the 78 *successful* paths). The last bar (right) is the mean number of actions required to find the target with a random policy. The graph indicates that the policy computed from the learned TPSR is close to optimal.

Finally, rather than computing $\widehat{P}_{\mathcal{T},ao,\mathcal{H}}$ directly, we instead compute $\widehat{U}^\top \widehat{P}_{\mathcal{T},ao,\mathcal{H}}$ for each pair $a, o$: the latter matrices are much smaller, and in our experiments, we saved substantially on both memory and runtime by avoiding construction of the larger matrices. To construct $\widehat{U}^\top \widehat{P}_{\mathcal{T},ao,\mathcal{H}}$, we restrict to those training trajectories in which the action at the middle time step (i.e., step 4) is $a$. Then, each element of $\widehat{P}_{\mathcal{T},ao,\mathcal{H}}$ is the empirical expectation (among the restricted set of trajectories) of the product of one indicative feature, one characteristic feature, and element $o$ of the observation kernel vector. So,

$$\widehat{U}^\top \widehat{P}_{\mathcal{T},ao,\mathcal{H}} = \frac{1}{w_a} \sum_{t=1}^{w_a} (\widehat{U}^\top \phi_t^{\mathcal{T}})(\phi_t^{\mathcal{H}})^\top \frac{1}{Z_t} K(o_t - o) \quad (14)$$

where $K(.)$ is the kernel function and $Z_t$ is the kernel normalization constant computed by summing over the 500 observation kernels for each $o_t$. Given the matrices $P_\mathcal{H}$, $P_{\mathcal{T},\mathcal{H}}$, and $P_{\mathcal{T},ao,\mathcal{H}}$, we can compute the TPSR parameters using the equations in Section 3.

### 5.3 Qualitative Evaluation

Having learned the parameters of the TPSR, the model can be used for prediction, filtering, and planning in the autonomous robot domain. We first evaluated the model *qualitatively* by projecting the sets of histories in the training data onto the learned TPSR state space: $\widehat{U}^\top \widehat{P}_\mathcal{H}$. We colored each datapoint according to the average of the red, green, and blue components of the highest probability observation following the projected history. The features of the low dimensional embedding clearly capture the topology of the major features of the robot's visual environment (Figure 1(C-D)), and continuous paths in the environment translate into continuous paths in the latent space (Figure 2(B)).

### 5.4 Planning in the Learned Model

To test the quality of the learned model, we set up a navigation problem where the robot was required to plan a set of actions in order to reach a goal image (looking directly at the blue wall). We specified a large reward (1000) for this observation, a reward of $-1$ for colliding with a wall, and 0 for every other observation. We next learned a reward function by linear regression from the histories embedded in the learned TPSR state space to the reward specified at each image that followed an embedded history. We used the reward function to compute an approximate value function using the Perseus algorithm with discount factor $\gamma = .8$, a prediction horizon of 10 steps, and with the 8000 embedded histories as the set of belief points. The learned value function is displayed in Figure 2(A). Once the approximate value function has been learned, and an initial belief specified, the robot greedily chooses the action which maximizes the expected value. The initial beliefs were computed by starting with $b_1$ and then incorporating 3 random action-observation pairs. Examples of paths planned in the learned model are presented in Figure 2(B); the same paths are shown in geometric space (recall that the robot only has access to images; the geometric space is *never* observed by the robot) in Figure 2(C). Note that there are a *set* of valid target positions in the environment since one can receive an identical close-up image of a blue wall from anywhere along the corresponding edge of the environment.

The reward function encouraged the robot to navigate to a specific set of points in the environment, therefore the planning problem can be viewed as solving a shortest path problem. Even though we don't encode this intuition into our algorithm, we can use it to quantitatively evaluate the performance of the policy in the original system. First we randomly sampled 100 initial histories in the environment and asked the robot to plan a path based on its learned policy. The robot was able to reach the goal in 78 of the trials. In 22 trials, the robot got stuck repeatedly taking alternating actions whose effects cancelled (for example, alternating between turning $-15°$ and $15°$).[5] When the robot *was* able to reach the goal, we compared the number of actions taken both to the *minimal path*, calculated by A* search in the robot's configuration space given the *true underlying position*, and to a random policy. Note that comparison to the optimal policy is somewhat unfair: in order to recover the optimal policy the robot would have to know its true underlying position (which is not available to it), our model assumptions would have to be exact, and the algorithm would need an unlimited amount of training data. The results, summarized in Figure 2(D), indicate that the TPSR policy is close to the optimal policy in the original system. We think that this result is remarkable, especially given that previous approaches have encountered significant difficulty modeling continuous domains [12] and domains with similarly high levels of complexity [33].

## 6. CONCLUSIONS

We have presented a novel *consistent* subspace identification algorithm that simultaneously solves the *discovery* and *learning* problems for TPSRs. In addition, we provided two extensions to the learning algorithm that are useful in practice, while maintaining consistency: characteristic and indicative features only require one to know relevant features of tests and histories, rather than sets of core tests and histories, while kernel density estimation can be used to find observable operators when observations are real-valued. We also showed how point-based approximate planning techniques can be used to solve the *planning* problem in the learned model. We demonstrated the representational capacity of our model and the effectiveness of our learning algorithm by learning a very compact model from simulated autonomous robot vision data. We closed the loop by successfully planning with the learned models, using Perseus to approximately compute the value function and optimal policy for a navigation task. To our knowledge this is the first instance of learning a model for a simulated robot in a partially observable environment using a consistent algorithm and successfully planning in the learned model. We compare the policy generated by our model to a bound on the best possible value, and determine that our policy is close to optimal.

We believe the spectral PSR learning algorithm presented here, and subspace identification procedures for learning PSRs in general, can increase the scope of planning under uncertainty for autonomous agents in previously intractable scenarios. We believe that this improvement is partly due to the greater representational power of the PSR as compared to POMDPs and partly due to the efficient and statistically consistent nature of the learning method.

## 7. REFERENCES

[1] K. J. Aström. Optimal control of Markov decision processes with incomplete state estimation. *Journal of Mathematical Analysis and Applications*, 10:174–205, 1965.

[2] J. Bilmes. A gentle tutorial on the EM algorithm and its application to parameter estimation for gaussian mixture and hidden markov models. Technical Report, ICSI-TR-97-021, 1997.

[3] M. Bowling, P. McCracken, M. James, J. Neufeld, and D. Wilkinson. Learning predictive state representations using non-blind policies. In *Proc. ICML*, 2006.

[4] A. R. Cassandra, L. P. Kaelbling, and M. R. Littman. Acting Optimally in Partially Observable Stochastic Domains. In *Proc. AAAI*, 1994.

[5] Eyal Even-Dar and Sham M. Kakade and Yishay Mansour. Planning in POMDPs Using Multiplicity Automata. In *UAI*, 2005.

[6] H. Jaeger, M. Zhao, A. Kolling. Efficient Training of OOMs. In *NIPS*, 2005.

[7] D. Hsu, S. Kakade, and T. Zhang. A spectral algorithm for learning hidden markov models. In *COLT*, 2009.

[8] M. T. Izadi and D. Precup. Point-based Planning for Predictive State Representations. In *Proc. Canadian AI*, 2008.

[9] H. Jaeger. Observable operator models for discrete stochastic time series. *Neural Computation*, 12:1371–1398, 2000.

[10] M. James and S. Singh. Learning and discovery predictive state representations in dynamical systems with reset. In *Proc. ICML*, 2004.

[11] M. R. James, T. Wessling, and N. A. Vlassis. Improving approximate value iteration using memories and predictive state representations. In *AAAI*, 2006.

[12] N. K. Jong and P. Stone. Towards Employing PSRs in a Continuous Domain. Technical Report UT-AI-TR-04-309, University of Texas at Austin, 2004.

[13] M. Littman, R. Sutton, and S. Singh. Predictive representations of state. In *Advances in Neural Information Processing Systems (NIPS)*, 2002.

[14] M. Zhao and H. Jaeger and M. Thon. A Bound on Modeling Error in Observable Operator Models and an Associated Learning Algorithm. *Neural Computation*.

[15] A. McCallum. Reinforcement Learning with Selective Perception and Hidden State. PhD Thesis, University of Rochester, 1995.

[16] P. McCracken and M. Bowling. Online discovery and learning of predictive state representations. In *Proc. NIPS*, 2005.

[17] J. Pineau, G. Gordon, and S. Thrun. Point-based value iteration: An anytime algorithm for POMDPs. In *Proc. IJCAI*, 2003.

[18] J. Pineau, G. Gordon, and S. Thrun. Anytime point-based approximations for large POMDPs.

---

[5]In an actual application, we believe that we could avoid getting stuck by performing a short lookahead or simply by randomizing our policy; for purposes of comparison, however, we report results for the greedy policy.

*Journal of Artificial Intelligence Research (JAIR)*, 27:335–380, 2006.

[19] M. Rosencrantz, G. J. Gordon, and S. Thrun. Learning low dimensional predictive representations. In *Proc. ICML*, 2004.

[20] S. Ross and J. Pineau. Model-Based Bayesian Reinforcement Learning in Large Structured Domains. In *Proc. UAI*, 2008.

[21] G. Shani, R. I. Brafman, and S. E. Shimony. Model-based online learning of POMDPs. In *Proc. ECML*, 2005.

[22] S. M. Siddiqi, B. Boots, and G. J. Gordon. Reduced-Rank Hidden Markov Models. *http://arxiv.org/abs/0910.0902*, 2009.

[23] B. W. Silverman. *Density Estimation for Statistics and Data Analysis*. Chapman & Hall, 1986.

[24] S. Singh, M. James, and M. Rudary. Predictive state representations: A new theory for modeling dynamical systems. In *Proc. UAI*, 2004.

[25] S. Singh, M. L. Littman, N. K. Jong, D. Pardoe, and P. Stone. Learning predictive state representations. In *Proc. ICML*, 2003.

[26] S. Soatto and A. Chiuso. Dynamic data factorization. Technical report, UCLA, 2001.

[27] E. J. Sondik. The Optimal Control of Partially Observable Markov Processes. PhD. Thesis, Stanford University, 1971.

[28] M. T. J. Spaan and N. Vlassis. Perseus: Randomized point-based value iteration for POMDPs. *Journal of Artificial Intelligence Research*, 24:195–220, 2005.

[29] P. Van Overschee and B. De Moor. *Subspace Identification for Linear Systems: Theory, Implementation, Applications*. Kluwer, 1996.

[30] E. Wiewiora. Learning predictive representations from a history. In *Proc. ICML*, 2005.

[31] D. Wingate. Exponential Family Predictive Representations of State. PhD Thesis, University of Michigan, 2008.

[32] D. Wingate and S. Singh. On discovery and learning of models with predictive representations of state for agents with continuous actions and observations. In *Proc. AAMAS*, 2007.

[33] D. Wingate and S. Singh. Efficiently learning linear-linear exponential family predictive representations of state. In *Proc. ICML*, 2008.

[34] B. Wolfe, M. James, and S. Singh. Learning predictive state representations in dynamical systems without reset. In *Proc. ICML*, 2005.